\documentclass[10pt, conference, compsocconf]{IEEEtran}
\usepackage{cite}
\usepackage{amsmath}
\usepackage{balance}
\usepackage{amssymb}
\usepackage[utf8]{inputenc}
\usepackage{color, soul}
\usepackage{url}
\usepackage{textcomp}
\usepackage{tabularx}
\usepackage{graphicx}
\usepackage[caption=false]{subfig}
\usepackage{wrapfig}

\def\BibTeX{{\rm B\kern-.05em{\sc i\kern-.025em b}\kern-.08em
    T\kern-.1667em\lower.7ex\hbox{E}\kern-.125emX}}

\ifCLASSINFOpdf
\else
\fi
\begin{document}
\IEEEoverridecommandlockouts
%
\title{Computational Histological Staining and Destaining of Prostate Core Biopsy RGB Images with Generative Adversarial Neural Networks}


\author{\IEEEauthorblockN{Aman Rana$^{1}$, Gregory Yaunery$^{1}$, Alarice Lowe$^{2}$, Pratik Shah$^{1\dagger}$ \thanks{$^{\dagger}$Corresponding author: \textit{pratiks@media.mit.edu}}} \\
\IEEEauthorblockA{$^{1}$MIT Media Lab, Massachusetts Institute of Technology, Cambridge, MA, USA \\ Email: \{arana,gyauney,pratiks\}@media.mit.edu}
\IEEEauthorblockA{$^{2}$Brigham and Women's Hospital, Harvard Medical School, Boston, MA, USA \\ Email: alowe@bwh.harvard.edu}
\thanks{Published at 2018 17th IEEE International Conference on Machine Learning and Applications}
\thanks{© 2018 IEEE.  Personal use of this material is permitted.  Permission from IEEE must be obtained for all other uses, in any current or future media, including reprinting/republishing this material for advertising or promotional purposes, creating new collective works, for resale or redistribution to servers or lists, or reuse of any copyrighted component of this work in other works.}
\thanks{DOI: 10.1109/ICMLA.2018.00133}
}


%
%

\maketitle

\begin{abstract}
Histopathology tissue samples are widely available in two states: paraffin-embedded unstained and non-paraffin-embedded stained whole slide RGB images (WSRI). Hematoxylin and eosin stain (H\&E) is one of the principal stains in histology but suffers from several shortcomings related to tissue preparation, staining protocols, slowness and human error. We report two novel approaches for training machine learning models for the computational H\&E staining and destaining of prostate core biopsy RGB images. The staining model uses a conditional generative adversarial network that learns hierarchical non-linear mappings between whole slide RGB image (WSRI) pairs of prostate core biopsy before and after H\&E staining. The trained staining model can then generate computationally H\&E-stained prostate core WSRIs using previously unseen non-stained biopsy images as input. The destaining model, by learning mappings between an H\&E stained WSRI and a non-stained WSRI of the same biopsy, can computationally destain previously unseen H\&E-stained images. Structural and anatomical details of prostate tissue and colors, shapes, geometries, locations of nuclei, stroma, vessels, glands and other cellular components were generated by both models with structural similarity indices of ~0.68 (staining) and ~0.84 (destaining). The proposed staining and destaining models can engender computational H\&E staining and destaining of WSRI biopsies without additional equipment and devices.
\end{abstract}


%
\IEEEpeerreviewmaketitle

\section{Introduction}
Histopathology involves the visual examination of the structure and morphology of a stained tissue section under a microscope by a pathologist for diagnosis of various abnormalities. A histopathological analysis following H\&E staining is considered the gold standard for the diagnosis for the majority of cancer types in liver, prostate, lung, kidney and other organs, and a variety of other diseases in humans and model systems used for biomedical research \cite{rubins_pathology}. H\&E stain consists of two components - hematoxylin dye that selectively stains the nuclei dark blue and eosin dye that stains the cytoplasm and stroma various shades of pink and the red blood cells dark red to facilitate vivid visualization and discernment of abnormalities in biopsy \cite{staining_protocol}. H\&E staining of tissue biopsies and subsequent visual examination by pathologists present several challenges such as variability and inconsistencies introduced by tissue preparation and staining protocols, human errors and also requires significant processing time and costs \cite{sources_of_variability}.  Whole slide imaging (WSI) is a method to capture super high-resolution RGB images of stained pathology slides up to 40x resolutions; provides a way to normalize stain variation and can provide valuable diagnostic value in digital format \cite{farahani2015whole,cho2017neural,zanjani2018stain,bug2017context,macenko2009method,mukhopadhyay2018whole}.

Previous reports describe approaches for recovering stained images of tissue biopsy using hyperspectral and multispectral imaging systems and multichannel image segmentation methods. Bautista \textit{et al.} compare linear and non-linear mappings of the spectral transmittance data between nonstained and H\&E-stained multispectral images resulting in computationally stained multi-spectral images that are then converted to RGB format \cite{bautista}. Bayramo\~glu \textit{et al.} use hyperspectral transmittance spectra of the nonstained images and the corresponding microscopy images of H\&E-stained slides to learn non-linear mappings between these image pairs using a conditional generative adversarial network (cGAN) \cite{bayramoglu}. Another study describes unsupervised segmentation for low-contrast multichannel color nonstained pathology images by non-linearly mapping such images to an increased number of channels using an empirical kernel map in combination with non-negative matrix factorization \cite{kopriva2015unsupervised}. Amrania \textit{et. al} propose an instrument that use a bespoke IR filters at different wavelengths to capture nuclei and cytoplasm in the tissue \cite{amrania2012digistain}. Others have reported using computational analysis and the non-linear mapping of spectral data from chemical imaging using infrared (IR) microscopy from non-stained images to stain cellular structures; and combining fluorescence and reflectance mosaics to generate virtual H\&E images \cite{mayerich2015stain,bini2011confocal}. However, all these methods suffer from significant limitations such as a) excitation by specific wavelengths (UV light) and acquisition of specialized hyperspectral, auto-fluorescence, fluorescence, multispectral images or repeated acquisition of IR spectra using expensive systems \cite{bautista,bayramoglu,kopriva2015unsupervised,mayerich2015stain,rivenson2018deep}; b) stain only a few cellular components with low accuracies and limited colors \cite{bautista}; c) loss of information in the stained images is quite significant precluding evaluation for diagnostic needs \cite{bayramoglu}; d) mappings are incapable of processing complex functions and high dimensional data to fully represent in the H\&E-stained images \cite{bautista,kopriva2015unsupervised,mayerich2015stain}. 

We tackle the problems of digital H\&E staining of nonstained paraffin-embedded WSRI to circumvent the manual staining process, and digital H\&E destaining of stained images. To our knowledge, we are the first to report digital staining and destaining of these existing slide images at the point-of-care without additional equipment or devices. We also devise a novel loss function that enforces tissue structure preservation in GAN outputs.

\begin{figure*}[ht]
	\subfloat[]{
    	\includegraphics[width=\textwidth]{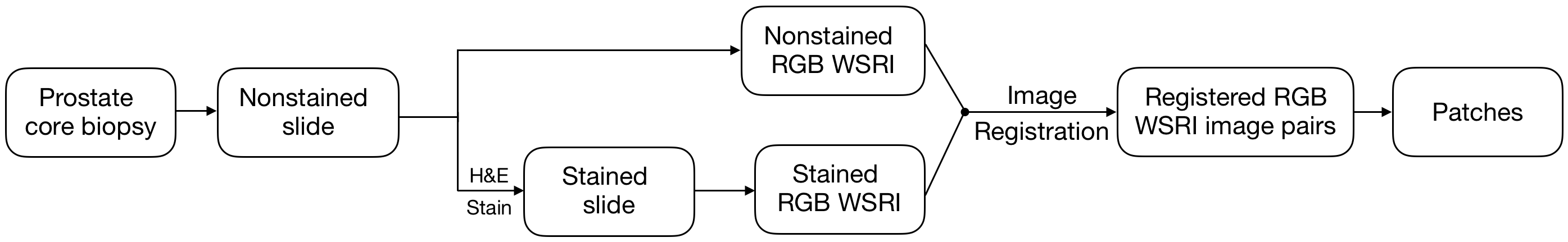}
        \label{fig: process_fig}
    } \\
    \subfloat[]{
    	\includegraphics[width=0.37\textwidth]{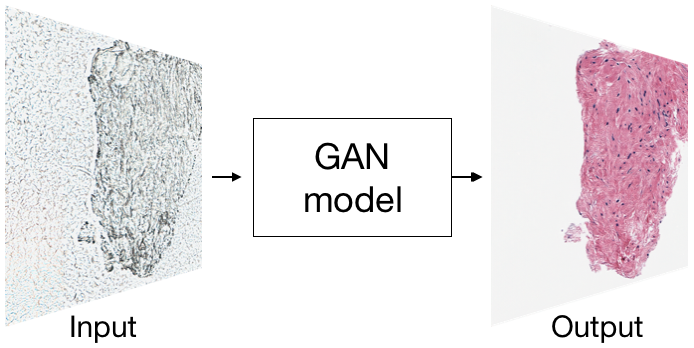}
        \label{fig: generator_input}
    }
    \hfill
    \vline
    \hfill
    \subfloat[]{
    	\includegraphics[width=0.57\textwidth]{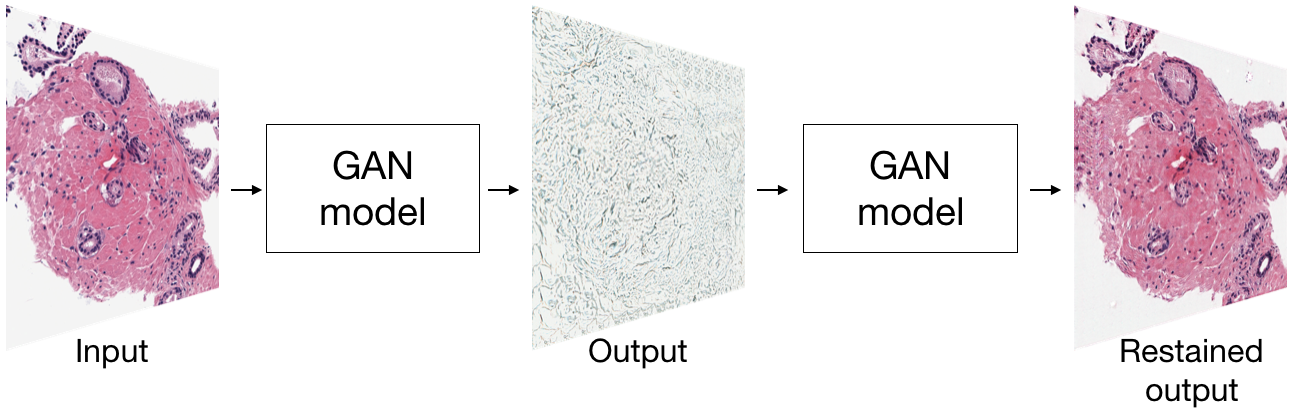}
        \label{fig: discriminator_input}
    }
  	\caption{Preprocessing flowchart and whole slide RGB images (WSRI) of prostate core biopsy used by generative machine learning models for computational H\&E staining and destaining. (a) Data acquisition, H\&E staining, registration and patch creation process for generating input images for machine learning models (b) Computational staining models use nonstained images (input) and generate predicted H\&E stained images (output); (c) The computational destaining model uses H\&E stained images (input) and generates predicted destained images (output) that were then validated using a secondary staining model (restained output)}
\end{figure*}

\section{Proposed methodology}
\subsection{Data collection and preprocessing}
The Partners Human Research Committee (Boston, MA) approved protocol 2014P002435, after which excess material from prostate core biopsies performed in the course of routine clinical care (2014-2017) at Brigham and Women's Hospital (BWH), Boston, MA, were obtained for this study. Briefly, prostate core biopsy specimens were immediately fixed in 10\% formalin, paraffin embedded, cut into 4-micron thick sections and placed on standard glass slides that were placed in archival storage at room temperature. Nonstained paraffin-embedded slides were scanned with the Aperio ScanScope XT system (Leica Biosystems, Buffalo Grove, IL) at 20$\times$ magnification. Subsequently, nonstained paraffin-embedded slides were stained with H\&E on the Agilent Dako Autostainer (Agilent, Santa Clara, CA), and these stained slides were re-scanned on the Aperio ScanScope XT at 20$\times$ magnification at the Harvard Medical School Tissue Microarray \& Imaging Core (TMIC). Nonstained-stained image pairs were registered using Adobe Photoshop (Adobe Systems, San Jose, CA).

The size of pathology WSRIs was too large (approx. 40,000$\times$40,000 pixels) to input directly into standard deep learning architectures.  We used a sliding window to extract 1024$\times$1024 pixel patches from each registered WSRI nonstained-stained pair. The sliding window stride was set at one-fourth the patch size resulting in 52,196 patch-pairs from 19 WSRI pairs. Images generated by computational staining and destaining machine learning models were compared to ground truth H\&E-stained images acquired from the TMIC by a structural similarity (SSIM) index and by an expert pathologist.

\subsection{Network architecture}
A generative adversarial network (GAN) is a type of deep learning architecture that consists of two network components: a generator $G$ that tries to generate realistic outputs from the given input and a discriminator $D$ that learns a binary classification to differentiate between the synthetic images produced by $G$ from the real images \cite{goodfellow2014generative} in the training dataset. A conditional GAN (cGAN) is a GAN where the output is additionally conditioned on an input image \cite{cGAN_paper}. cGANs are well suited for generative tasks for images and photographs \cite{bayramoglu,cGAN_paper,neffgenerative}.

Computational staining and destaining of patches extracted from nonstained and H\&E-stained WSRI was performed using a novel loss function for the cGAN architecture \cite{cGAN_paper}. The U-net architecture was used to create the generator while PatchGAN was used for the discriminator. The pix2pix architecture was updated with the following changes: layers were added to both the encoder and decoder part of the cGAN U-net model to facilitate input images of 1024$\times$1024 pixels; and the loss function was modified for better results. The cGAN network with the default loss function, GAN loss + L1 loss, generated computationally stained images that contained high-level tiling noise (data not shown). Multiple regularization terms were tested for increased preservation of structural information in the generated images. Pearson's correlation coefficient (CC) was chosen as it reduced high level tiling artifacts in the generated images. Pearson's correlation coefficient term was calculated between the output and target image. The overall loss function with the Pearson\'s correlation coefficient regularization term is:

\begin{align*}
\mathcal{L}_{cGAN}(G, D) &=\mathbb{E}_{x,y}[log D(x, y)]\ + \\ & \ \ \ \ \alpha \mathbb{E}_{x,z}[log(1-D(x,\ G(x, z)))]\\
\mathcal{L}_{L1}(G) &= \mathbb{E}_{x,y,z}(\parallel y-G(x,z) \parallel_{1})\\
\mathcal{L}_{correlation\ coeff}(G) &= \mathbb{E}_{x,y,z}(corr\_coef(y,\ G(x,z)))
\label{equation_1}
\end{align*}

Our final loss function is:
\begin{align*}
G^{*} =\ & arg\ \underset{G}{min}\ \underset{D}{max}\ \mathcal{L}_{cGAN}(G, D)\ + \\ & \lambda \mathcal{L}_{L1}(G) + \gamma \mathcal{L}_{correlation\ coeff}(G)
\end{align*}

where $x$ is the input image, $y$ is the target image and $z$ is the random noise, added as dropout in our work. $\mathcal{L}_{cGAN}(G, D)$ is the cGAN loss function, $\mathcal{L}_{L1}(G)$ is the L1 loss between the output of the generator and the target image, and $\mathcal{L}_{correlation\ coeff}(G)$ is the proposed term that calculated the Pearson\'s correlation coefficient between the generator output and target image. $\alpha=1$, $\lambda=100$ and $\gamma=10$ gave best results.

\begin{figure*}[h]
	\begin{center}
  		\includegraphics[width=\textwidth]{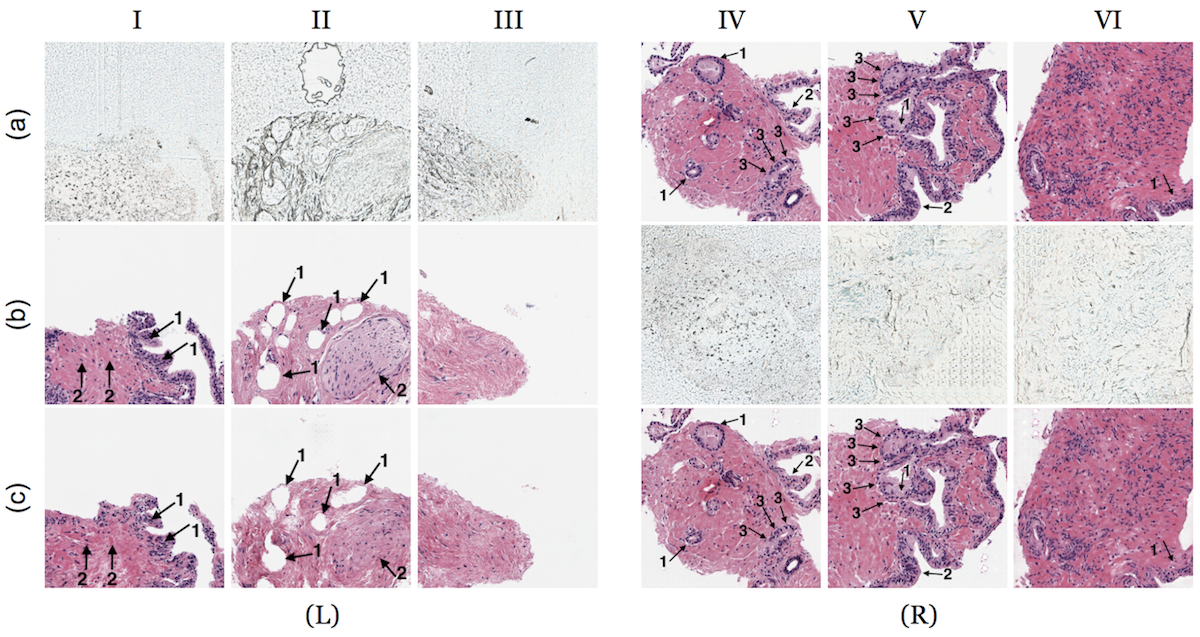}
	\end{center}
	\caption{Representative input and output images from computational staining and destaining networks.  Images in panel L show input in row (a), target (ground truth) in row (b) and output generated by the computational staining network in row (c).  Images in panel R show input in row (a), computational destaining network generated output in row (b) and output generated by the secondary staining network in row (c). Arrows represent microscopic morphological and structural features of prostrate core biopsy samples.}
    \label{fig: model_results}
\end{figure*}

\subsection{Training}
Two separate machine learning models were trained: 1) a staining model that generates H\&E-stained prostate core WSRI using previously unseen paraffin embedded non-stained RGB WSRI as input, and 2) a destaining model that reverses the process and computationally destains previously unseen H\&E-stained RGB images. Both models were trained using 40,148 patches from 14 WSRIs and validated on 12,048 patches from 5 WSRIs. Images used for training and those used for validating performance did not overlap. The discriminator was trained for every generator training step. Both networks were trained for 10 epochs each using Adam optimization and a batch size of one on a TITAN X GPU (NVIDIA, Santa Clara, CA) with 12 GB of VRAM and CUDA acceleration to speed up training. One epoch of training (40,148 training patches) took approximately 10 GPU hours. The patches were randomly flipped and dropout was used to prevent over-fitting.

\section{Results}

\subsection{Computational H\&E staining}
Results from all staining and destaining networks are shown in Figure 2. Output images from the computational staining network [Figure 2 (L) (c); Supplementary figure \ref{supp_fig: staining_grid}] accurately predict the spatial location and silhouette of the tissue. The computationally stained output images were compared to corresponding target TMIC H\&E-stained images using SSIM (0.6760) and CC (0.6878).


Examination by an expert pathologist showed that the computational staining network predicts the presence of different histological structures and cell types such as prostatic glands, prostatic stroma, nerve, adipocytes and vascular spaces, but rarely predicts a structure that does not exist (arrows in Figure 2 (L) (c); Supplementary figure \ref{supp_fig: staining_grid}). The morphology of the prostatic stroma is replicated best. The cellular detail of the prostatic gland epithelium is not represented well, with loss of cell polarity, nuclear location, and cytoplasmic features. Some structures, such as adipocytes and nerve, are suggested (arrows in Figure 2 (L) (c); Supplementary figure \ref{supp_fig: staining_grid}).

In a computationally stained patch shown in Figure 2 (c) I, the presence and morphologic appearance of prostatic glands (arrows labeled 1) and stroma (arrows labeled 2) are depicted as the increased cellular density identified at the periphery of the tissue composed of epithelioid cells and relative hypocellularity in the center composed of spindle cells with indistinct cytoplasmic edges, respectively. The palisaded nature of the prostatic gland nuclei and the distinct edges present on the prostatic gland cytoplasm can be improved (Figure 2 (c) I). In Figure 2 (c) II, the location and presence of adipocytes (arrows labeled 1) and nerve (arrows labeled 2) are represented accurately. 

\begin{table}
    \caption{Performance of the secondary staining network using outputs from the computational destaining network; as compared to actual H\&E stained images.}
    \begin{tabular*}{\linewidth}{l @{\extracolsep{\fill}}r}
        \hline
        \textbf{Parameter} & \textbf{Destaining} \\
        & \textbf{network} \\
        \hline
        SSIM 					& 0.84 \\
        Correlation coefficient & 0.897 \\
        \hline
    \end{tabular*}
    \label{table: destaining_model_results}
\end{table}

\subsection{Computational destaining of H\&E images}
The output from the computational destaining network [Figure 2 (R) (b); Supplementary figure \ref{supp_fig: destaining_grid}] shows that the trained model can successfully destain H\&E stained RGB images. We validated accuracy of the destaining network by restaining the generated destained computational images by a secondary H\&E staining model and comparing to ground truth TMIC stained images [Figure 1 (c)]. Output images from the secondary network showed high similarities to the ground truth TMIC H\&E-stained images [Figure 2 (R) (c); Supplementary figure \ref{supp_fig: destaining_grid}] (Table \ref{table: destaining_model_results}). The destaining model identified the location and silhouette of all tissue present on the slides; its representation of morphologic features was also accurate. The presence of a clear basement membrane interface between glands and stroma, prostatic epithelium nuclear location, and cytoplasmic membrane detail are present in nearly all images shown in Figure 2 (R) (c) and supplementary figure \ref{supp_fig: destaining_grid}.

In individual computationally restained patches shown in Figure 2 (c) (IV and V), all structures are easily identifiable morphologically, including improved nuclear polarity/location (arrows labeled 1), crisper cytoplasmic borders (arrows labeled 2), and a more identifiable basement membrane interface between glands and stroma (arrows labeled 3). Rare structures, such as the glandular epithelium (bottom right in Figure 2 (c) VI arrow labeled 1), were not distinctly identifiable in the test images.

\section{Discussion and future work}
In this study, we report fully trained cGAN computational staining and destaining models, which learn highly non-linear mappings between high resolution nonstained and H\&E-stained RGB image pairs of prostate core biopsy tissue samples. The computational staining model predicted location and silhouette of the prostate tissue, different histological structures/cell types such as prostatic glands, prostatic stroma, nerve, adipocytes and vascular spaces and associated colors in our validation images with good accuracies.

Bautista \textit{et al.} attempt to classify cellular components (like nuclei, cytoplasm and red blood cells) and white spaces using multispectral nonstained and H\&E-stained images. The classification is discrete and uses biopsy samples from serial sections of the same tissue \cite{bautista}. On the other hand, our method is able to classify a gamut of color intensities and uses nonstained and stained whole slide RGB images from the same tissue biopsy slide for each patient. Kopriva \textit{et al.} produce a segmentation of low-contrast multi-channel nonstained images, whereas our method predicts the RGB intensities of H\&E-stained WSRIs without specifying a segmentation. The segmentations produced by the method described in Kopriva \textit{et al.} visually correspond to the ground truth but show imprecise classification for cellular structures \cite{kopriva2015unsupervised}. Our method's prediction of the visual qualities of H\&E-stained images allows the predicted images to be used for additional interpretation by medical experts and segmentation. Mayerich \textit{et al.} train a two layer artificial neural network to learn a mapping from FT-IR spectroscopy nonstained image pixels to the corresponding RGB color intensity, but it does not accurately recover spatial resolution and necessitates custom and variable methods of sample preparation and requires specialized imaging systems which are neither standardized nor readily accessible in pathology laboratories \cite{mayerich2015stain}. Bayramo\~glu use the transmittance spectra of hyperspectral nonstained images and corresponding microscopy images of size 1000$\times$1000 px and train a cGAN architecture to report generated H\&E-stained lung biopsy images. The generated images are low resolution and suffer from information loss (0.38 SSIM index) \cite{bayramoglu}.

Detailed expert pathologist analyses outline successes and challenges in staining of paraffin-embedded WSRI microscopic structures that quantitative image-based metrics like SSIM and correlation coefficient may not address \cite{pambrun2015limitations}. Input image pairs (nonstained and H\&E stained) used for training in our work may have differences in field of view, illumination and focal planes resulting in decrease in registration and mapping accuracies for the computational staining and destaining models. The suitability of the generated images for tumor diagnosis has not been evaluated. We are actively investigating methods to control for these factors to achieve higher accuracies. Because of a lack of discernible features and associated translucency in paraffin-embedded nonstained biopsy images \ref{supp_fig: destaining_grid}, a secondary H\&E staining model was used to check the accuracy of the destaining network. The destaining network performed with higher accuracies than our staining model. Higher accuracies observed in the results from the destaining network could be attributed to the greater amounts of structural information that is preserved from the input stained images. In summary, the models described in this paper do not require hyperspectral or multispectral imaging systems or additional biochemical steps. The proposed models preserve microscopic morphological and sub-cellular structures in the H\&E stained and destained images of paraffin embedded prostrate core biopsy samples making them practical. To our knowledge, this is the first study describing the computational staining and destaining of RGB images of whole slide prostate core biopsy sections using cGAN.

\renewcommand{\thefigure}{A.\arabic{figure}}
\setcounter{figure}{0}

\section{Supplementary material}

\subsection{Computational staining model results}
Supplementary figure \ref{supp_fig: staining_grid} shows computationally H\&E stained output patches and corresponding input images and ground truth patches.

\subsection{Computational destaining model results}
Supplementary figure \ref{supp_fig: destaining_grid} shows computationally H\&E destained output patches; corresponding computationally restained patches and input and ground truth patches.

\balance
\bibliographystyle{unsrt}
\bibliography{bibliography}

\onecolumn

\begin{figure}
  \begin{center}
    \includegraphics[width=0.9\linewidth]{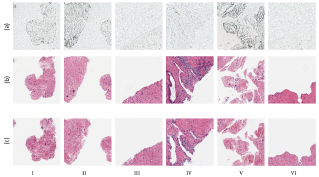}
    \caption{Input image patches (a), ground truth image patches (b) and computationally H\&E stained output image patches (c)}
    \label{supp_fig: staining_grid}
  \end{center}
\end{figure}

\begin{figure}
  \begin{center}
    \includegraphics[width=0.90\linewidth]{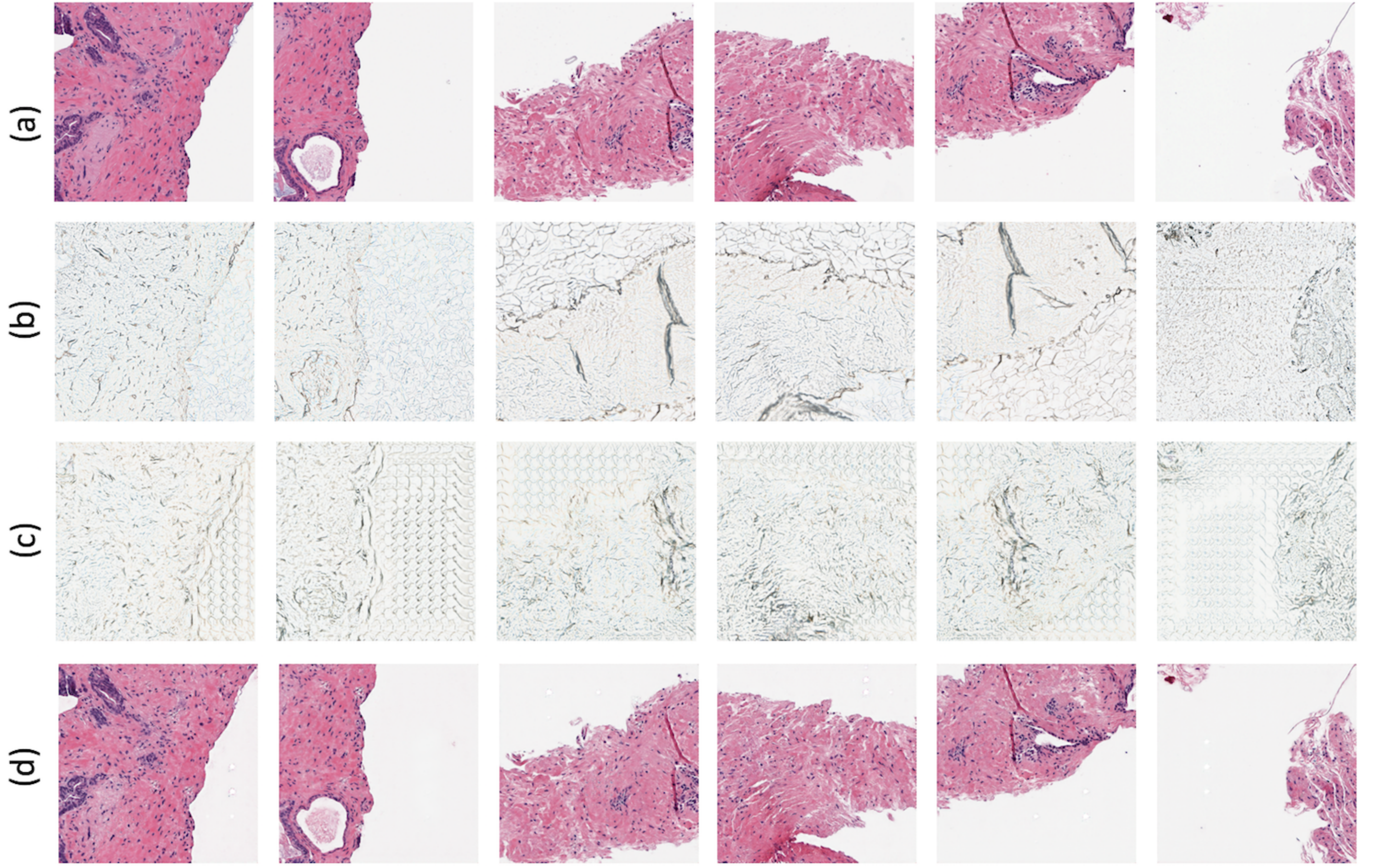}
    \caption{H\&E stained image patches (a), non-stained image patches (b), computationally destained image patches (c), computationally restained output image patches (d)}
    \label{supp_fig: destaining_grid}
  \end{center}
\end{figure}

\end{document}